\documentclass[runningheads]{llncs}
\usepackage[T1]{fontenc}
\usepackage{amssymb}
\usepackage{longtable}
\usepackage{CJK}
\usepackage{color}
\usepackage{algorithm}
\usepackage{algpseudocode}
\usepackage{authblk}
\usepackage{tablefootnote}

\usepackage{graphicx}
\begin{document}
\title{Can ChatGPT Replace Traditional KBQA Models? An In-depth Analysis of the Question Answering Performance of the GPT LLM Family}%\thanks{Supported by organization x.}
\titlerunning{Analysis of the QA Performance of the GPT LLM Family.}
%\orcidID{0000-1111-2222-3333}

\author{Yiming Tan \inst{1,4\star} 
\and Dehai Min \inst{2,4\star} 
\and Yu Li\inst{2,4} 
\and Wenbo Li\inst{3}  
\and Nan Hu\inst{2,4} 
\and Yongrui Chen\inst{2,4}  
\and  Guilin Qi \inst{2,4} \footnote[2]{Corresponding author}
}
\renewcommand{\thefootnote} {\fnsymbol{footnote}}
\footnotetext[1]{Yiming Tan and Dehai Min contribute equally to this work.}
\authorrunning{Y. Tan et al.}


\institute{School of Cyber Science and Engineering, Southeast University, Nanjing, China 
\and
School of Computer Science and Engineering, Southeast University, Nanjing, China 
\and School of Computer Science and Technology, Anhui Unviersity, Hefei, China 
\and Key Laboratory of New Generation Artificial Intelligence Technology and Its Interdisciplinary Applications (Southeast University), Ministry of Education, China
\email{\{tt\underline{ }yymm,zhishanq,yuli\underline{ }11,nanhu,yrchen,gqi\}@seu.edu.cn},  \email{wenboli@stu.ahu.edu.cn} \\
 }
 
\maketitle              % typeset the header of the contribution

\begin{abstract} 
ChatGPT is a powerful large language model (LLM) that covers knowledge resources such as Wikipedia and supports natural language question answering using its own knowledge. Therefore, there is growing interest in exploring whether ChatGPT can replace traditional knowledge-based question answering (KBQA) models. Although there have been some works analyzing the question answering performance of ChatGPT, there is still a lack of large-scale, comprehensive testing of various types of complex questions to analyze the limitations of the model. In this paper, we present a framework that follows the black-box testing specifications of CheckList proposed by \cite{ribeiro2020beyond}. We evaluate ChatGPT and its family of LLMs on eight real-world KB-based complex question answering datasets, which include six English datasets and two multilingual datasets. The total number of test cases is approximately 190,000. In addition to the GPT family of LLMs, we also evaluate the well-known FLAN-T5 to identify commonalities between the GPT family and other LLMs. The dataset and code are available at https://github.com/tan92hl/Complex-Question-Answering-Evaluation-o
f-GPT-family.git

\end{abstract}
 
\section{Introduction}
Given its extensive coverage of knowledge from Wikipedia as training data and its impressive natural language understanding ability, ChatGPT has demonstrated powerful question-answering abilities by leveraging its own knowledge. Additionally, a study conducted by \cite{petroni2019language} suggests that language models can be considered as knowledge bases (KBs) to support downstream natural language processing (NLP) tasks. This has led to growing interest in exploring whether ChatGPT and related large language models (LLMs) can replace traditional Knowledge-Based Question Answering (KBQA) models.

There have been many evaluations of ChatGPT \cite{zhong2023can,kocon2023chatgpt,chen2023robust,zhuo2023exploring,huang2023chatgpt,omar2023chatgpt,wang2023can,wang2023cross,qin2023chatgpt,bang2023multitask}, some of which include the testing of question answering tasks and have yielded interesting conclusions: 
for example, \cite{omar2023chatgpt}  showed that ChatGPT has lower stability than traditional KBQA models on a test set of 200 questions, and \cite{bang2023multitask} found that ChatGPT is a "lazy reasoner" that suffers more with induction after analyzing 30 samples. However, due to the limited number of test cases, it is difficult to perform a comprehensive evaluation of ChatGPT’s performance on the KBQA task based on these findings. Moreover, the reliability of these findings still requires further testing for validation. We find that the difficulty in answer evaluation is the main reason why existing works have not conducted large-scale KBQA tests on ChatGPT, which outputs sentences or paragraphs that contain answers rather than an exact answer. Furthermore, due to the influence of the generated textual context, the answer sequence of ChatGPT may not necessarily correspond strictly to entity names in the knowledge base. Therefore, the traditional Exact Match (EM) metric cannot directly evaluate the output of ChatGPT for question-answering. Consequently, most of the works mentioned above rely on manual evaluation.

In this paper, we select the KB-based Complex Question Answering (KB-based CQA) task to comprehensively evaluate the ability of LLMs to answer complex questions based on their own knowledge. This task requires the model to use compositional reasoning to obtain the answer to the question, which includes multi-hop reasoning, attribute comparison, set operations, and other complex reasoning. We believe that evaluating ChatGPT’s performance in complex knowledge question answering using its own knowledge can help us understand whether existing LLMs have the potential to surpass traditional KBQA models or whether ChatGPT is already capable of replacing the current best KBQA models. Therefore, we collect test data from existing KB-based CQA datasets and establish an evaluation framework.

Our evaluation framework consists of two parts: 1) the feature-driven unified labeling method is established for the KBQA datasets involved in the testing; and 2) the evaluation of answers generated by LLMs. Inspired by the approach of using multiple scenario tags to evaluate language models in the HELM framework \cite{liang2022holistic}, we label each test question with unified answer-type, inference-type, and language-type tags. In the answer evaluation part, we first improve the Exact Match (EM) method so that it can be used to evaluate the accuracy of LLMs' output. The main process of improved EM is to extract potential answer phrases from the LLM output through constituent trees as the candidate answer pool, and then match them with the reference answer pool formed by annotated answers and aliases provided by wikidata.
Next, we follow the CheckList testing specification \cite{ribeiro2020beyond} and set up three tests: the minimal functionality test (MFT), invariance test (INV) \cite{segura2016survey}, and directional expectation test (DIR). Along with an overall evaluation, these tests assess the LLMs' capability, stability, and control when answering questions and performing specific reasoning operations.

Finally, we collect six English real-world KB-based CQA datasets and two multilingual real-world KB-based CQA datasets for our evaluation experiment, with a scale of approximately 190,000 questions, including approximately 12,000 multilingual questions covering 13 languages. 
In the experiment, we mainly compare the QA performance differences between the traditional the current state-of-the-art (SOTA) models and the GPT family models \cite{brown2020language,ouyang2022training,openai2023gpt4}. In addition, we also introduce the open-source LLM FLAN-T5 \cite{chung2022scaling} model as a representative of the non-GPT family for comparison. Like ChatGPT, all the LLMs involved in the comparison in this paper use their own knowledge to answer questions and are considered unsupervised models.

Our key findings and insights are summarized as follows:

ChatGPT and the LLMs of GPT family outperform the best traditional models on some old datasets like WQSP and LC-quad2.0, but they still lag behind the current state-of-the-art on the latest released KBQA datase such as KQApro and GrailQA. 

GPT family LLMs and the FLAN-T5 model tend to have similar tendencies in terms of strengths and weaknesses when answering different types of questions.

Using chain-of-thought prompts in CheckList testing enhances GPT LLMs' ability to answer specific questions but may negatively impact other question types, suggesting their potential and sensitivities for future task-specific applications.

\section{Related Work}
\subsection{Large language models and prompting}
In recent years, LLMs and prompt learning have attracted considerable attention. Groundbreaking studies such as \cite{petroni2019language,jiang2020can,brown2020language} revealed that LLMs, when given appropriate textual prompts, can perform a wide range of NLP tasks with zero-shot or few-shot learning without gradient updates.
% In this paradigm, the language capabilities of LLMs and effective prompts complement each other. 
On the one hand, improved prompting can enable the information contained in the LLM to be more accurately applied to the target task, and early representative works include \cite{reynolds2021prompt,qin2021learning}
% , which redefine downstream tasks through prompts. 
The chain-of-thought (CoT) \cite{wei2022chain} method is a distinguished approach in effective prompt research. 
% By inducing LLMs to generate intermediate reasoning steps before generating answers, 
CoT enables LLMs to have a better understanding and think more when answering questions. 
On the other hand, much work has been done to improve the natural language understanding ability of LLMs, including Gopher \cite{Rae2021scaling} and PaLM \cite{chowdhery2022palm}, which aim to extend LLMs. 
Undoubtedly, ChatGPT has garnered significant attention as a prominent LLM due to its remarkable natural language understanding abilities. It is trained on the GPT-3.5 series of models \cite{fu2022does} using RLHF.

\subsection{Evaluation of the large language model}
While LLMs have demonstrated outstanding natural language understanding and generation capabilities, it is still necessary to further research their strengths, limitations, and potential risks to fully understand their advantages. Recently, many works aimed at evaluating LLMs have been proposed \cite{chang2023survey}, including general benchmarks like HELM \cite{liang2022holistic}, Bigbench \cite{srivastava2022beyond}, Promptbench \cite{zhu2023promptbench}, and MME \cite{fu2023mme}. These aim to categorize and summarize multiple existing tasks, providing a macro-level assessment of LLM performance and potential biases. Other studies focus on specific NLP tasks, such as summarization \cite{bang2023multitask}, question-answering \cite{bang2023multitask,bai2023benchmarking,omar2023chatgpt}, and machine translation \cite{lyu2023new}. In these existing works, the advantages of the general benchmark approaches lie in their fine-grained sample classification and high testing efficiency. However, these benchmarks are limited by the use of automated metrics, which restrict the diversity of testing objectives. On the other hand, evaluating task-specialized LLMs introduces more manually defined testing objectives, such as interpretability, determinism, robustness, and question understanding. Nevertheless, due to manual testing costs, these evaluations often rely on small samples (less than 10k) and coarsely categorized datasets.

In this paper, we combine the strengths of both benchmark studies and task-specific manual evaluations to test the GPT family LLMs. To achieve this, we adopt a strategy inspired by HELM \cite{liang2022holistic}, which uses multiple feature labels to describe and categorize task types, especially complex problem types. Additionally, we incorporate the manually predefined testing objectives from \cite{omar2023chatgpt} and combine them with the CheckList natural language model's black-box testing strategy. This comprehensive and diverse testing approach allows us to draw more comprehensive and valuable conclusions.

\subsection{Black-box testing of the NLP model}
%CheckList
The prohibitive expense associated with training LLMs renders white-box testing an impractical approach. Consequently, the majority of assessment efforts presently concentrate on black-box evaluation approaches for LLMs. For example, the methods used by \cite{belinkov2019analysis,rychalska2019models} for evaluating robustness, the methods used by \cite{wu2019errudite} for adversarial changes, and attention and interpretability within LLMs research conducted by \cite{wang2019superglue}. The most comprehensive approach currently available is the CheckList approach proposed by \cite{ribeiro2020beyond}, which categorizes evaluation targets into three parts: the minimum functionality test (MFT), invariance test (INV), and directional expectation test (DIR). The MFT examines a model’s basic functionality, INV examines whether the model can maintain functional correctness when non-answer-affecting information is added to the input, and DIR examines whether the model can output the expected result when the input is modified. In this work, we follow the idea of CheckList and use CoT prompting to generate test cases for DIR.

\begin{figure}[t]\centering
\includegraphics[width=0.92\textwidth]{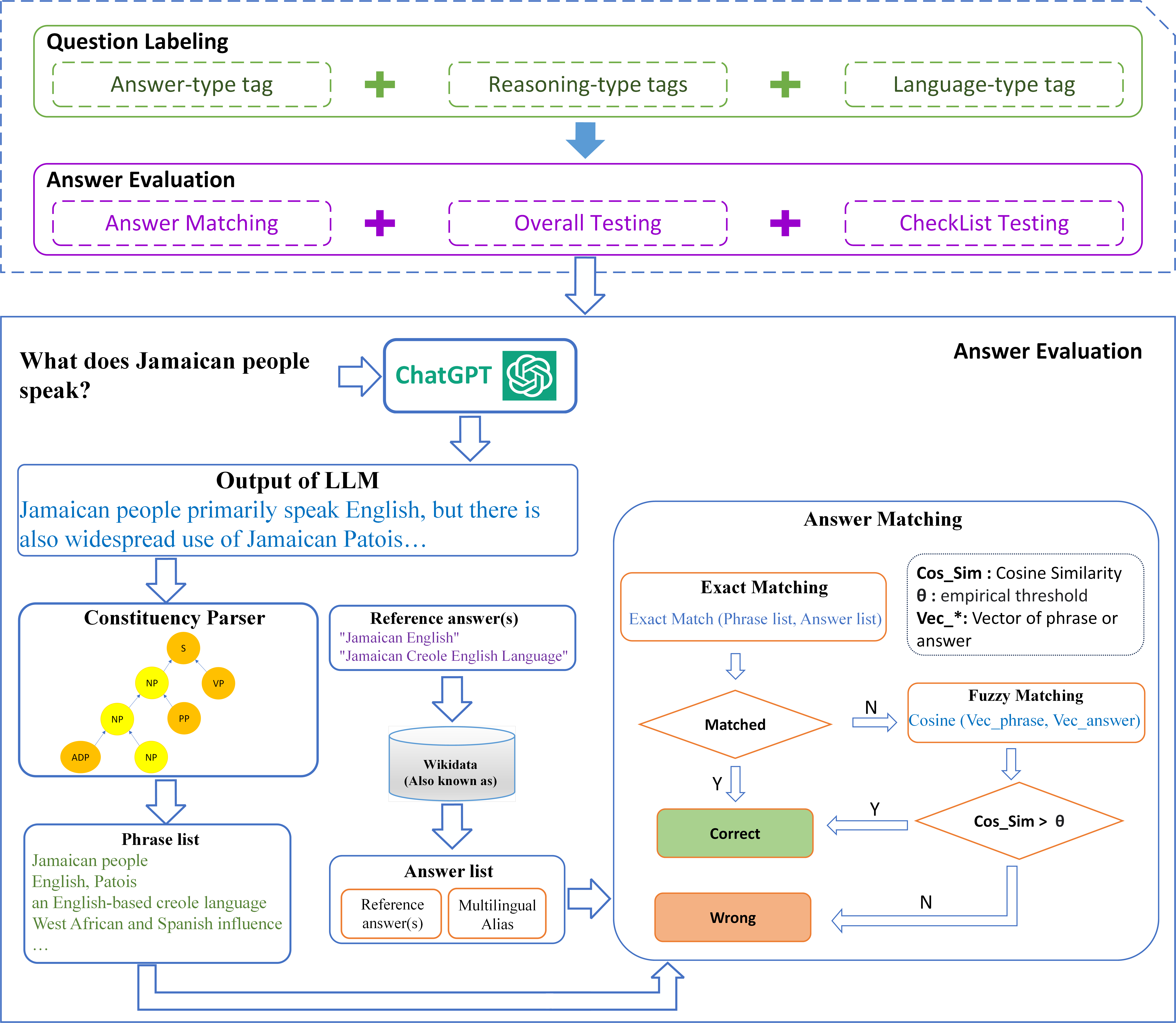}
\caption{Overview of proposed Evaluation Framework.} \label{fig1}
\end{figure}

\begin{table}[t]\centering
\caption{The feature-driven question tags defined in this paper.}\label{tab11}
\setlength{\tabcolsep}{1.5mm}{
    \begin{tabular}{lp{8cm}}
    \hline
     Answer type &  Description  \\
    \hline
     MISC & The answer to the question is the miscellaneous fact defined by the named entity recognition task. \\
     PER &  The answer to the question is the name of a person.\\
     LOC &  The answer to the question is a location. \\
     WHY &  The answer explains the reasons for the facts mentioned in the question. \\
     DATE & The answer to the question is a date or time.  \\
     NUM &  The answer to the question is a number. \\
     Boolean & The answer to the question is yes or no.  \\
     ORG & The answer to the question is the name of a organization.  \\
     UNA &  The input question is unable to answer.  \\
    \hline
    Reasoning type &  Description\\
    \hline
    SetOperation & The process of obtaining answers involves set operations.\\
    Filter & The answer is obtained through condition filtering.\\
    Counting & The process of obtaining an answer involves counting operations. \\
    Comparative & The answer needs to be obtained by comparing or sorting numerical values.\\
    Single-hop & Answering questions requires a single-hop Reasoning.\\
    Multi-hop & Answering questions requires multi-hop Reasoning.\\
    Star-shape & The reasoning graph corresponding to inputting question is star-shape.\\
    \hline
    \end{tabular}
    }
\end{table}

\begin{figure}[h]
\includegraphics[width=\textwidth]{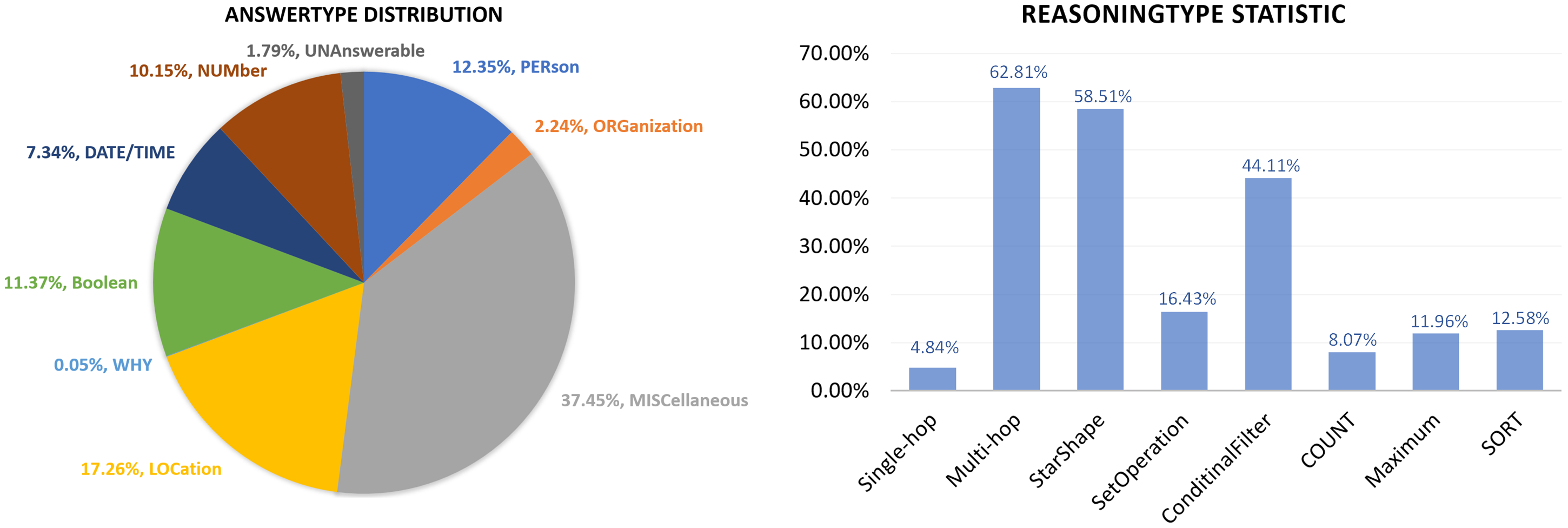}
\caption{The distribution of feature labels in the collect KB-based CQA datasets} \label{fig2}
\end{figure}

\section{Evaluation Framework}
As mentioned in Section 1, our KBQA evaluation framework consists of two parts. The first part aims to assign uniform feature labels to the questions in the datasets. The second part includes an improved Exact Match answer evaluation strategy and an extended CheckList test. Figure \ref{fig1} illustrates the overall process of the framework. The detailed process is described in the following section.

% \begin{figure}[t]
% \includegraphics[width=\textwidth]{ans_matching.png}
% \caption{The answer matching strategy via phrase extraction, alias collection and cosine similarity threshold.} \label{fig3}
% \end{figure}

\subsection{Feature-driven unified question labeling}
We collect multiple existing KB-based CQA datasets for the evaluation. However, due to the different annotation rules used for features such as answer and reasoning type in each dataset, we need to establish a standardized and unified set of question feature tags for evaluating and analyzing question types.

Referring to the question tags provided by existing KBQA datasets \cite{ngomo20189th,longpre2021mkqa,cao2022kqa,yih2016value}, we categorize the tags that describe the features of complex questions into three types, including answer type, reasoning type and language type. Table \ref{tab11} lists the eight answer type tags and seven reasoning type tags we defined. Generally, a question contains one answer type tag, one language type tag and several reasoning type tags. Figure \ref{fig2} presents the label distribution of the data collected in this paper. For an input question, our labeling process is as follows: when the dataset provides question type tags, we simply match them to our feature tag list. When no tag is provided, we use an existing bert-base-NER model \cite{ner2003,kenton2019bert} to identify the type of answer, and use keywords in SPARQL to identify the type of inference.

\subsection{Answer evaluation}
% \subsubsection{Answer matching strategy}
The output of traditional KBQA models typically takes two forms: either a SPARQL query or a precise answer. The evaluation strategy for traditional KBQA models is based on exact match (EM), which involves comparing the model's output with a reference answer or to assess its accuracy. 
However, without adding additional prompts, LLMs generate text paragraphs containing answers, rather than precise answers. Furthermore, this answer may be a restatement of the reference answer.

\textbf{Extended Answer Matching} To obtain evaluation results on KBQA outputs of LLMs resembling exact match, we propose an extended answer matching approach. This approach consists of three main parts:
1) Parsing LLMs' output using constituent trees \cite{he-choi-2021-stem} to extract NP or VP root node phrases as the candidate answer pool.
2) Expanding each reference answer using multilingual alias lists from Wikidata, including various names and aliases.
3) Using m-bert \cite{kenton2019bert} to calculate the maximum Cosine similarity between reference and candidate answers for precise match evaluation, with a fuzzy matching strategy applied only to non-"NUM, DATE, Boolean" answer types.
% To obtain evaluation results on KBQA outputs of LLMs that closely resemble exact match, we propose an extended answer matching approach. As Figure \ref{fig1} shows, this approach consist of three main parts: 1) We parse the output of LLMs using constituent trees \cite{he-choi-2021-stem} and extract all subtrees corresponding to phrases with root nodes as NP or VP. We then treat these extracted phrases as the candidate answer pool. 2) We expand each individual reference answer to the reference answer pool using the multilingual alias lists provided by Wikidata. This pool includes reference answers, their various names, and their multilingual aliases. 3) To reduce the potential false negative issues in Exact Match, we use m-bert \cite{kenton2019bert} to obtain and calculate the maximum Cosine similarity between reference and candidate answers when a precise match is not achieved. This helps us determine the correctness of the answers and design an algorithm to obtain an empirical threshold. It is important to note that this fuzzy matching strategy is only employed when the answer type is not "NUM, DATE, Boolean”.

%新增图片
\begin{figure}[t]
\includegraphics[width=\textwidth]{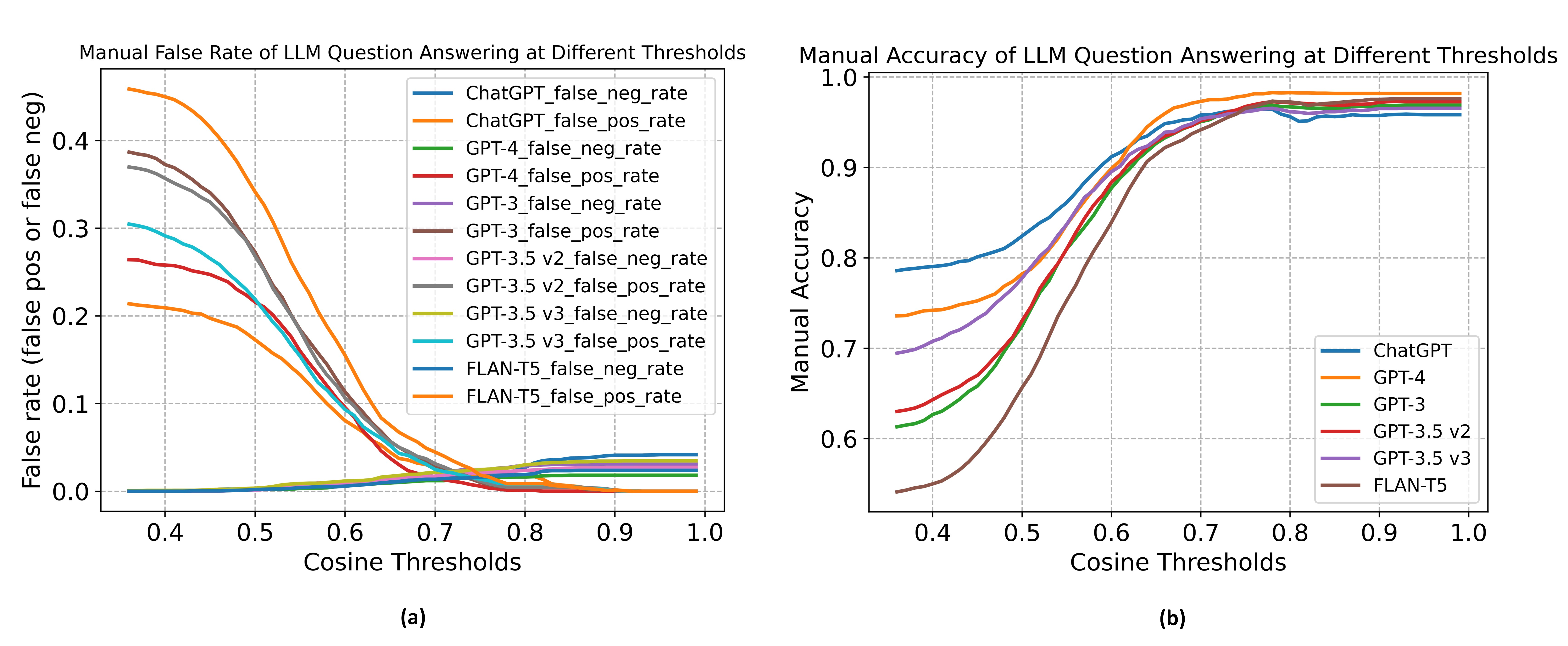}
\caption{(a) The GPT family and T5 models show changing error rates on sampled questions as the threshold varies. (b) LLMs' QA accuracy (evaluated manually) on sampled questions varies with the threshold.} \label{fig10}
\end{figure}

\textbf{Threshold Selection and Sensitivity Analysis} 
As shown in Figure \ref{fig10} (a), the analysis of various models reveals that using only EM evaluation for answers (threshold=1) may result in 2.38\%-4.17\% (average 3.89\%) false negative cases. To address this issue, we opt for establishing a fuzzy matching process based on cosine similarity to alleviate the problem. However, selecting an inadequate threshold may introduce additional false positive issues. Therefore, we followed the steps below to find an empirical threshold that minimizes the overall false rate (false pos + false neg) across all models:
(1) We randomly sampled 3000 question samples from the test data of answer types involved in fuzzy matching and manually verified the correctness of the six LLM output answers shown in Figure \ref{fig10}(a) (binary labels, correct/incorrect).
(2) We calculate the minimum cosine similarity (the value is 0.38) between the gold answer and its aliases, and used it as the lower bound for finding the threshold.
(3) We observed the changes in false rates for each model as the threshold increased from 0.38 to 1 and selected the threshold of 0.78 that minimized the average false rate across models. From the Figure \ref{fig10}(a), it can be observed that the false rates of each model stabilize around this value. 
To evaluate the sensitivity of model performance to the threshold, as shown in Figure \ref{fig10}(b), we compared the accuracy of each model on the test data as the threshold varied. The accuracy of each model tended to stabilize when the threshold was >0.7.
Finally, we use 0.78 as the empirical threshold for further experiments. Sampling tests show that this threshold decreases the average false rate from 3.89\% to 2.71\%.

\begin{figure}[t] \centering
\includegraphics[width=0.8\textwidth]{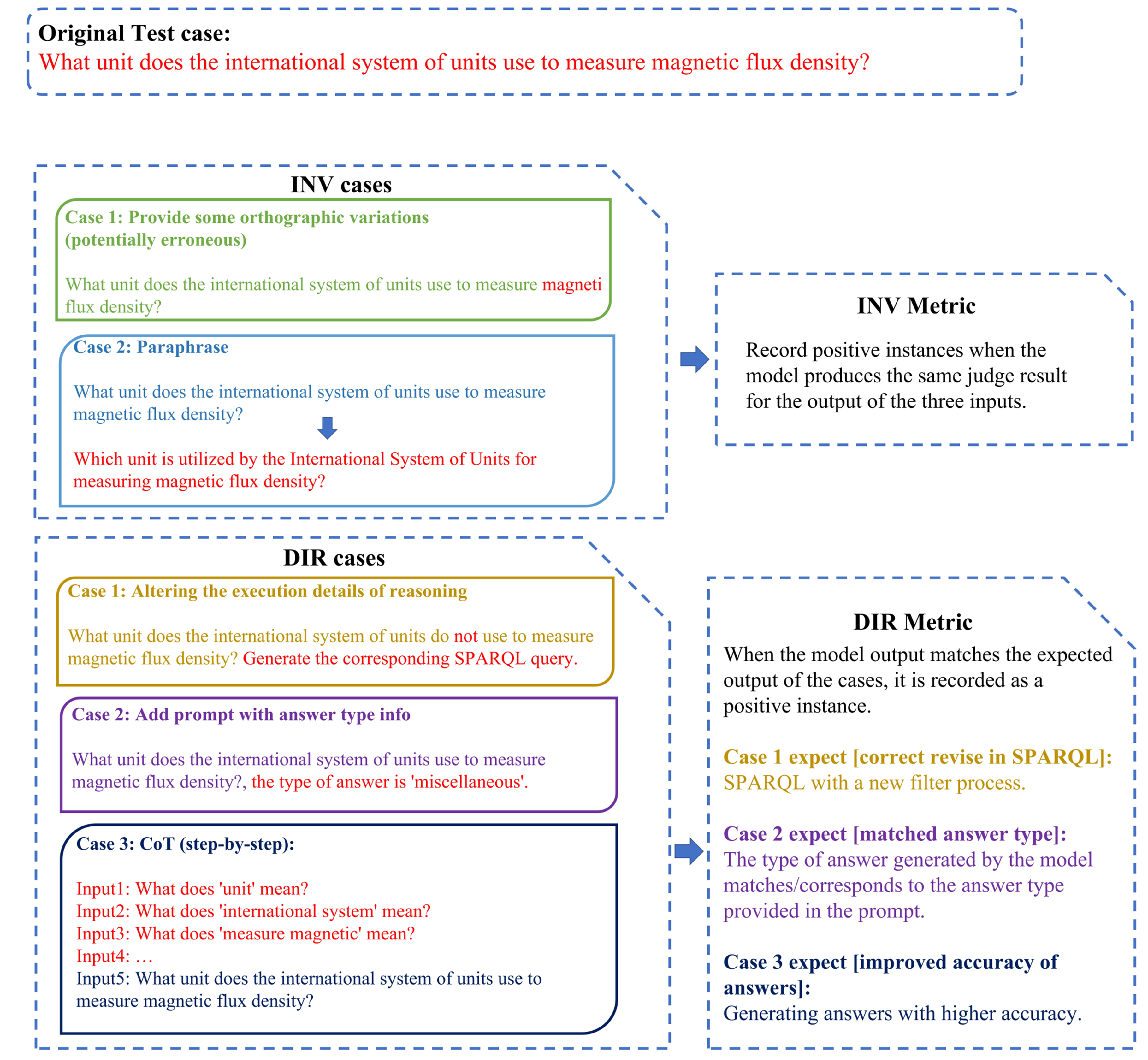}
\caption{Test cases design for INV and DIR.} \label{fig4}
\end{figure}

\subsection{CheckList testing}
Following the idea of CheckList, we also evaluate ChatGPT and other LLMs with three distinct objectives: (1) to evaluate the ability of LLMs to handle each feature in KB-based CQA through the MFT; (2) to evaluate the robustness of LLMs’ ability to handle various features in KB-based CQA scenarios through the INV; and (3) to evaluate whether the outputs of LLMs meet human expectations for modified inputs through the DIR, the controllability. 
The specific INV and DIR procedures are presented as follows, and Figure \ref{fig4} presents the instances:

\textbf{Minimum Functionality Test} In this work, we choose to examine the performance of LLMs in performing basic reasoning tasks by only including quesitons that involve a single type of reasoning operation. We then compare and analyze the performance differences of the models in answering questions that require performing a single reasoning operation versus those that require performing multiple reasoning operations.

\textbf{Invariance Test} We designe two methods to generate test cases for INV: the first method is to randomly introduce spelling errors into the original sentence, and the second method is to generate a question that is semantically equivalent (paraphrased) to the original sentence. Subsequently, we evaluate the invariance of the LLMs by checking the consistency of their correctness in the outputs generated from three inputs, i.e. the original test sentence, the version of the question with added spelling errors, and the paraphrased question.

\textbf{Directional Expectation Test}
In this study, we designed three modes for DIR test cases: (1) Replacing phrases related to reasoning operations in questions, to observe if LLMs' outputs correspond to our modifications. (2) Adding prompts with answer types after the original question text to check LLMs' ability to control the output answer type. (3) Using multi-round questioning inspired by CoT, where LLMs consider information related to key nouns before asking the original question, to observe the effectiveness and sensitivity of CoT prompts for different question types.
% In this study, we designed three modes to create DIR test cases. First, we replaced the phrases related to reasoning operations in the questions, demanding the model to output the corresponding SPARQL query statement, in order to observe whether the reasoning operations in the LLMs' outputs corresponded to our modifications. Second, we added prompts containing the answer type after the original question text to check whether LLMs can control the output answer type based on the prompts. Third, inspired by CoT, we used multi-round questioning to let LLMs first consider information related to key nouns in the question, and then ask the original question, so that LLMs can obtain the answer through a step-by-step thinking process, to observe the effectiveness and sensitivity of LLMs to CoT prompts for different types of questions.

\section{Experiments}
\subsection{Datasets}
To highlight the complexity of the testing questions and the breadth of the testing dataset, after careful consideration, we selected six representative English monolingual KBQA datasets and two multilingual KBQA datasets for evaluation. These datasets include classic datasets such as WebQuestionSP \cite{yih2016value}, ComplexWebQuestions \cite{talmor2018web}, GraphQ \cite{su2016generating} and QALD-9 \cite{ngomo20189th}, as well as newly proposed datasets such as KQApro \cite{cao2022kqa},  GrailQA \cite{gu2021beyond} and MKQA \cite{longpre2021mkqa}. Due to the limitations of the OpenAI API, we sampled some datasets, such as MKQA (sampled by answer type) and GrailQA (only using the test set). The collection size for each dataset and the scale we collected are summarized in Table \ref{tab1}.
% Due to OpenAI’s API limitations, we sample some datasets, such as MKQA (sampled by answer type) and GrailQA (test set only). The collection size of each dataset is summarized in Table \ref{tab1}.

\begin{table}[t]\centering
\caption{The Statistical of collected KB-based CQA datasets, "Col. Size" represents the size of the dataset we collected in our experiments. "Size" denotes the original size of the dataset.}\label{tab1}
% \resizebox{\linewidth}{!}{
\setlength{\tabcolsep}{1.8mm}{
    \begin{tabular}{lllc}
    \hline
    Datasets   & Size & Col. Size & Lang    \\
    \hline
    KQApro     & 117,970 & 106,173 & EN  \\%& 16,960 & 1,209 (363 pred, 846 attr)  \\
    LC-quad2.0 & 26,975 & 26,975 & EN  \\
    WQSP       & 4737 & 4,700 & EN  \\
    CWQ        & 31,158 & 31,158 & EN   \\
    GrailQA    & 64,331 & 6,763 & EN   \\
    GraphQ    & 4,776 & 4,776 & EN    \\
    \hline
    QALD-9     & 6,045 & 6,045 & Mul   \\
    MKQA       & 260,000 & 6,144 & Mul   \\
    \hline
    Total Collected      & & 194,782 &   \\
    \hline
    \end{tabular}
    }
\end{table}

\subsection{Comparative models}
\textbf{State-of-the-art models for each dataset} 
% We introduce current SOTA models' report scores from the KGQA leaderboard \cite{perevalov2022knowledge} for each dataset as comparison traditional KGQA models in this paper. This primarily reflects the comparison between ChatGPT and the latest traditional KBQA models on the overall results.
We introduce current SOTA models’ report scores from the KBQA leaderboard \cite{perevalov2022knowledge} for each dataset as traditional KBQA models in this paper for comparison. This primarily reflects the comparison between LLMs and traditional KBQA models in terms of the overall results.

\textbf{Large-language models of the GPT family} 
ChatGPT is a landmark model in the GPT family, and we believe that comparing it to its predecessors and subsequent versions is very valuable. By doing so, we can observe and analyze the technical increments of the GPT family at each stage and the benefits they bring. In this paper, we compare the GPT family models, which include GPT-3, GPT-3.5 v2, GPT-3.5 v3, ChatGPT (Their names on OpenAI's Model Index document are: text-davinci-001, text-davinci-002, text-davinci-003, gpt-3.5-turbo-0301) and the newest addition, GPT-4 \cite{openai2023gpt4}.

\textbf{Large-language model not belongs to GPT family} 
The LLM we have chosen is the famous FLAN-T5 (Text-to-Text Transfer Transformer 11B, [7]), which does not belong to the GPT family. Considering its multilingual question-answering ability and open-source nature, we have chosen it to participate in the comparison in this paper. FLAN-T5 is an encoder-decoder transformer language model that is trained on a filtered variant of CommonCrawl (C4) \cite{raffel2020exploring}. The release date and model size for this model are also based on \cite{raffel2020exploring}.

\subsection{Overall results}
The overall results are presented in Table \ref{tab2}. First, ChatGPT outperforms the current SOTA traditional models on three of the eight test sets, and the subsequently released GPT-4 surpasses on four test sets. By comparing the performance of GPT-4 and SOTA models, we can see that as LLMs represented by the GPT family, their zero-shot ability is constantly approaching and even surpassing traditional deep learning and knowledge representation models.

\begin{figure}[t]
\includegraphics[width=\textwidth]{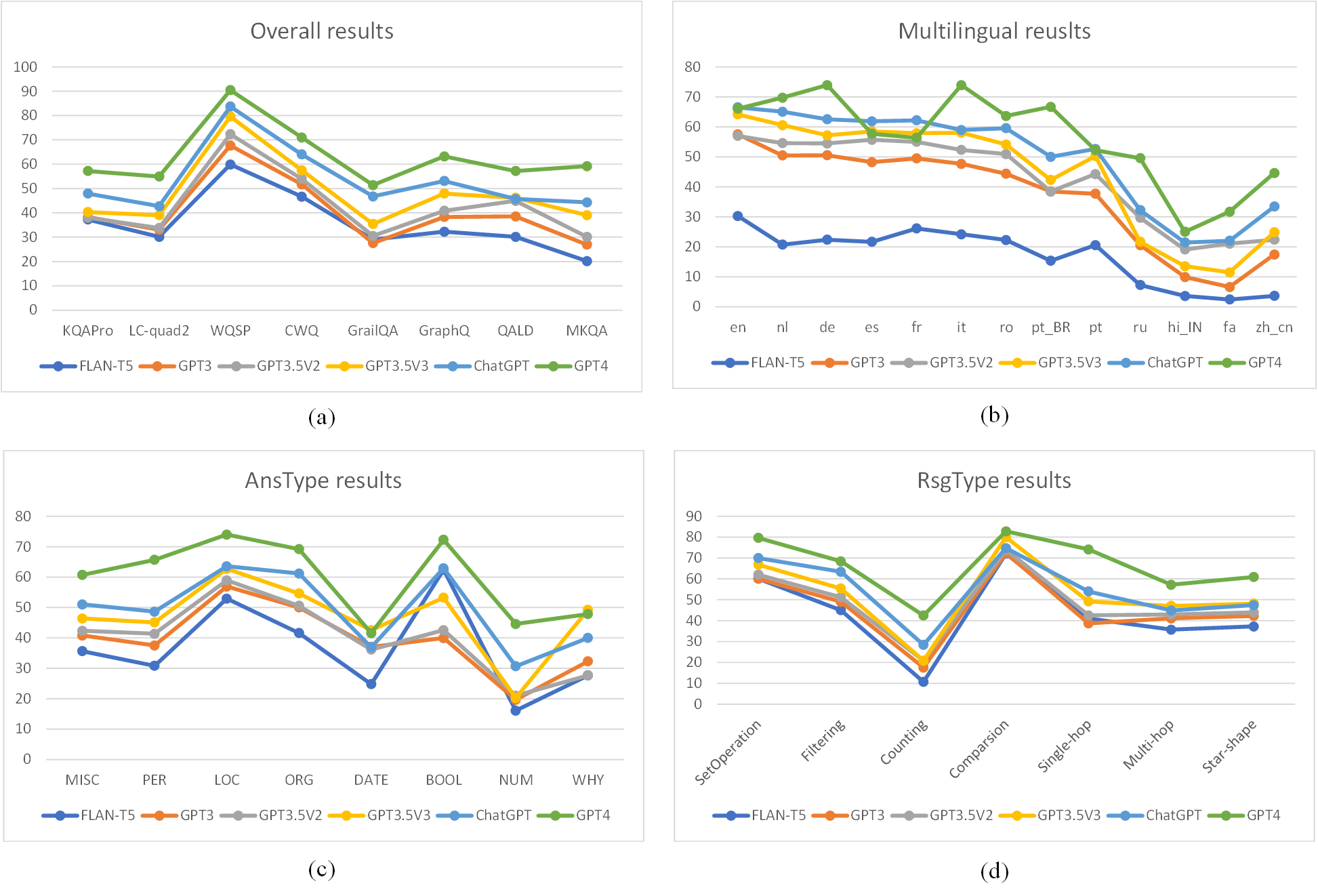}
\caption{(a) is the line chart based on Table \ref{tab2}, showing the EM scores of each model on different datasets. (b) corresponds to Table \ref{tab3}, with lines representing the EM scores of each model in different languages. (c) and (d) correspond to Table \ref{tab4}, reflecting the trend of EM scores of each model on different types of questions.} \label{fig5}
\end{figure}

Second, comparing models in the GPT family, the newer models perform better than the previous ones, as expected. Interestingly, the performance improvement of the new GPT models is relatively consistent across all datasets, as shown in Figure \ref{fig5}(a), where the line shapes of all GPT models are almost identical. This means that each generation of GPT models retains some commonalities. Based on the known cases, these commonalities may come from the transformer-based encoding. We will discuss in detail the impact they have in section 4.5. In addition, we can observe that the newer versions of the GPT model show increasingly significant improvements compared to the previous generations. 
% To be more specific, compared to GPT-3, the average improvement of GPT-3.5 v2 is 7.27\%, while GPT-3.5 v3 achieves a 13\% improvement. ChatGPT has reached an amazing 27.83\% improvement, thus gaining more attention. GPT-4 has achieved a 33.98\% improvement over GPT-3.5 v3, and a 19.31\% improvement over ChatGPT. 
%The development of LLMs, represented by GPT, is in a period of rapid growth, and its evolution is expected to be promising for a long time in the future.

% Based on these trends and improvement rates, we can make the following assumptions:

% (1) Each version iteration of the GPT family maintains a line shape that is almost identical to the others in Figure \ref{fig5}(a). We speculate that this is because the improvement of GPT models is not specific to downstream tasks but rather a more fundamental advancement. When we observe the improvement from the perspective of a single pipeline, this enhancement affects the most basic part of the pipeline, which may be a more reasonable natural language representation, more accurate natural language understanding, or better keyword recognition.

% (2) From GPT-3 to GPT-3.5 and GPT-4, the improvement rate between each version iteration is increasing, indicating that LLMs have significant room for improvement in future versions. As a model that was released during the same period as GPT-3.5v3, ChatGPT is a more special model, with a much higher improvement rate than GPT-3.5v3. Comparing the technical differences between the two, this gain comes from learning dialog.

Third, as shown in Figure \ref{fig5}(a), although FLAN-T5’s overall performance is weaker than that of the GPT family, its line shape is quite similar to that of the GPT family. This further supports our inference that the transformer-based architecture leads to commonalities in the abilities of current LLMs.
% Based on this, we infer that despite the significant performance gap, these transformer-based LLMs still share some commonalities, even including GPT-4, which incorporates multimodal data. Thus, we can anticipate that future LLMs may still maintain these commonalities as long as the underlying data encoding logic remains unchanged.

\begin{table}[t]\centering
\caption{Overall results of the evaluation. We compare the exact match of ChatGPT with current SOTA traditional KBQA models (fine-tuned (FT) and zero-shot (ZS)), GPT family LLMs, and Non-GPT LLM. In GraphQ, QALD-9 and LC-quad2, the evaluation metric used is F1, while other datasets use Accuracy (Exact match).}\label{tab2}. 
\resizebox{\linewidth}{!}{
\setlength{\tabcolsep}{0.4mm}{
    \begin{tabular}{l|l|l|l|l|l|l|l|l}
    \hline
    Datasets  & KQApro & LC-quad2 & WQSP & CWQ & GrailQA & GraphQ & QALD-9 & MKQA \\
    \hline
              & Acc  & F1 & Acc  & Acc &  Acc  & F1  & F1 & Acc \\
    \hline
     SOTA(FT)   & \textbf{93.85} \cite{perevalov2022knowledge}  &  33.10 \cite{pramanik2021uniqorn} & 73.10 \cite{hu2022logical}  & \textbf{72.20} \cite{hu2022logical}  & \textbf{76.31} \tablefootnote{https://dki-lab.github.io/GrailQA/} & 31.8 \cite{gu2022arcaneqa} & \textbf{67.82} \cite{purkayastha2022deep} & 46.00 \cite{longpre2021mkqa} \\
     SOTA(ZS) & 94.20 \cite{nie2022graphq} & -    & 62.98 \cite{ye2022rng} & -  & -  & -  & -  & - \\
     FLAN-T5    & 37.27 &  30.14 & 59.87  & 46.69  & 29.02 & 32.27 & 30.17 & 20.17 \\
     GPT-3      & 38.28 &  33.04 & 67.68  & 51.77  & 27.58 & 38.32 & 38.54 & 26.97 \\
     GPT-3.5v2  & 38.01 &  33.77 & 72.34  & 53.96  & 30.50 & 40.85 & 44.96 & 30.14 \\
     GPT-3.5v3  & 40.35 &  39.04 & 79.60  & 57.54  & 35.43 & 47.95 & 46.19 & 39.05 \\
     ChatGPT    & 47.93 & 42.76  & 83.70  & 64.02  & 46.77 & 53.10 & 45.71 & 44.30 \\
     GPT-4      & 57.20 & \textbf{54.95} & \textbf{90.45}  & 71.00 & 51.40 & \textbf{63.20} & 57.20 & \textbf{59.20} \\
    \hline
    \end{tabular}
    }}
\end{table}

\begin{table}[t] \centering
\caption{Comparison of LLMs on multilingual test sets.}\label{tab4}
\resizebox{\linewidth}{!}{
\setlength{\tabcolsep}{1.5mm}{
    \begin{tabular}{c|c|cccccc|cc}
    \hline
    Languages &  FLAN-T5 & GPT-3 & GPT-3.5v2 & GPT-3.5v3 & ChatGPT & GPT-4\\
    \hline
    en  & 30.29 & 57.53  & 56.99  & 64.16  & \textbf{66.49} & 66.09\\
    nl  & 20.75 & 50.47 & 54.58 & 60.56 & 65.05 & \textbf{69.72}\\
    de  & 22.40 & 50.54 & 54.48 & 57.17 & 62.54 & \textbf{73.91}\\
    es  & 21.68 & 48.22 & 55.70 & 58.50  & \textbf{61.87} & 57.69\\
    fr  & 26.16 & 49.46 & 55.02 & 57.89 & \textbf{62.19} & 62.00\\
    it  & 24.19 & 47.67 & 52.33 & 58.06  & 58.96 & \textbf{73.91}\\
    ro  & 22.28 & 44.38 & 50.94 & 54.12 & 59.55 & \textbf{63.41}\\
    pt\underline{ }br & 15.38  & 38.46 & 38.46 & 42.31 & 50.00 & \textbf{66.67}\\
    pt  & 20.58 & 37.70 & 44.26 & 50.27 & \textbf{52.64} & 52.25\\
    ru  & 7.29 & 20.58 & 29.69 & 21.68 & 32.24 & \textbf{49.58}\\
    hi\underline{ }in  & 3.61 & 9.93 & 19.13 & 13.54  & 21.48 & \textbf{25.00}\\
    fa & 2.45 & 6.59 & 21.09 & 11.49 & 22.03 & \textbf{31.71}\\
    zh\underline{ }cn & 3.65 & 17.45 & 22.40 & 24.87 & 33.46 & \textbf{44.62}\\
    \hline
    \end{tabular}
    }
    }
\end{table}

\subsection{Multilingual KBQA results}
Based on the results from MKQA and QALD-9, we further present the performance of LLMs on multilingual QA in Table \ref{tab4}. Despite the overall trend showing improvement in the model's ability to answer questions in different languages as the GPT family continues to iterate, we observe that GPT-4 has not surpassed ChatGPT in the four languages. This suggests that the evolution of GPT's multilingual capabilities may be starting to slow down.
Figure \ref{fig5}(b) shows the line chart of the EM scores of the models in each language. We can find that the shapes of the lines for GPT-3 and ChatGPT are very similar, while there is a significant change in the shape of the line for GPT-4. We believe that the main reason for this change is due to the introduction of multimodal data in GPT-4, which plays a positive role in mapping between some languages.

% The performance of the GPT family models in answering multilingual questions tends to improve with each version iteration. ChatGPT and GPT-4 are still the top-performing models, which is consistent with the overall results. 

% We can find that the GPT-3, GPT-3.5, and ChatGPT models maintain the same line shape as they update their versions, further supporting our previous conjecture that the improvement of GPT models lies in the non-task-specific part. This also indicates that, from the perspective of natural language processing, although GPT models are constantly updating, their proficiency in parsing languages has not undergone significant changes. This phenomenon will be a crucial reference for the applicability of the GPT family in multilingual tasks. On the other hand, the line shape of GPT-4 in Figure \ref{fig5}(b) shows a significant change compared to the GPT family, implying that the addition of multimodal information does not always have a positive effect on multilingual parsing.
% The line shape of FLAN-T5 in Figure \ref{fig5}(b) shows that it has similar proficiency in processing languages as the GPT family, further supporting our discussion on the "commonalities" among transformer-based LLMs in the previous section.

\begin{table}[t] \centering
\caption{Exact Match comparison based on Answer Types (AnsType) and Reasoning Types (RsgType) }\label{tab3}
% \resizebox{\linewidth}{!}{
\setlength{\tabcolsep}{1.5mm}{
    \begin{tabular}{c|c|ccc|cc}
    \hline
    MF &  FLAN-T5& GPT-3 & GPT-3.5v2 & GPT-3.5v3 & ChatGPT & GPT-4\\
    \hline
    \multicolumn{7}{c}{AnsType} \\
    \hline
    MISC    & 35.67  & 40.79  & 42.35  & 46.42  & 51.02 & \textbf{60.73}\\
    PER     & 30.84  & 37.53  & 41.36  & 45.10  & 48.65 & \textbf{65.71}\\
    LOC     & 52.91  & 56.92  & 58.93  & 62.71  & 63.55 & \textbf{73.98}\\
    ORG     & 41.62  & 50.01  & 50.58  & 54.62  & 61.18 & \textbf{69.20}\\
    DATE    & 24.81  & 37.07  & 36.15  & \textbf{42.54}  & 36.92 & 41.57\\
    Boolean & 62.43  & 39.96  & 42.56  & 53.23  & 62.92 & \textbf{72.28}\\
    NUM     & 16.08  & 19.66  & 21.01  & 20.31  & 30.70 & \textbf{44.59}\\
    WHY     & 27.69  & 32.31  & 27.69  & \textbf{49.23}  & 40.00 & 47.83\\
    UNA     &  - &  - &  - & -  & - & -\\
    \hline
    \multicolumn{7}{c}{RsgType} \\
    \hline
    SetOperation    & 60.11  & 60.12  & 62.03  & 66.86  & 70.00 & \textbf{79.70}\\
    Filtering   & 45.01  & 49.06 & 51.24  & 55.43  & 63.40 & \textbf{68.40}\\
    Counting    & 10.68  & 17.56  & 20.83 & 20.83  & 28.41 & \textbf{42.50}\\
    Comparison  & 72.13  & 72.44  & 74.00 & 80.00  & 74.74 & \textbf{82.79}\\
    Single-hop  & 41.00  & 38.72  & 42.54 & 49.22  & 54.00 & \textbf{74.14}\\
    Multi-hop   & 35.68  & 41.09  & 42.98 & 47.06  & 44.88 & \textbf{57.20}\\
    Star-shape  & 37.23  & 42.28  & 43.96 & 48.17  & 47.43 & \textbf{60.91}\\
    \hline
    \end{tabular}
    }
\end{table}

\subsection{Feature tags based results}
The results in Table \ref{tab3} show the performance of ChatGPT and other LLMs when answering different types of questions. As such, traditional models are not compared in this section. 
Overall, models from the GPT family are better at answering questions with boolean (yes/no) answers, questions about organizations and locations, as well as those involving set operations and numerical comparisons. However, they do not perform well when answering questions that require precise dates or involve numerical calculations. From the performance of models in the GPT family and Flan-T5, it can be found that Flan-T5 performs worse in all cases except for questions with boolean answer types. This is consistent with the conclusion of \cite{liang2022holistic}: The performance of knowledge-intensive tasks is closely related to the size of the model. For comparisons within the GPT family of models, following the iteration process of the GPT model summarized in \cite{fu2022does}, we also observe some positive effects of certain technical introductions on the model, including: 
% (1) Both GPT-3.5 v3 and ChatGPT are two different variants of models that use instruction tuning with RLHF. GPT-3.5 v3 restores some of the lost in-context learning abilities in GPT-3.5 v2 and further improves zero-shot capabilities, while ChatGPT seems to sacrifice some in-context learning abilities in exchange for modeling dialogue history. As a result, GPT-3.5 v3 performs better in answering multi-hop and star-shape questions that require distinguishing entity mentions through context. 
(1) \cite{fu2022does} point out that GPT-3.5 v3 has better in-context learning abilities, while ChatGPT sacrifices these abilities in order to model dialogue history. This may explain why GPT-3.5 v3 performs better in answering multi-hop and star-shaped questions that require distinguishing entity mentions through context. 
(2) ChatGPT's dialogue learning helps it better answer short questions (Single-hop). 
(3) The GPT-3.5 v2, obtained through language model and code training followed by supervised instruction tuning, but its overall capabilities do not appear to have significantly improved compared to GPT-3. The possible reason could be that alignment harms performance, and the alignment tax offsets the increase in zero-shot ability obtained through training \cite{ouyang2022training,liang2022holistic}. 
(4) One possible reason why the successor models outperform GPT3.5 V2 in most aspects is that the complex reasoning ability acquired through training on code, which did not manifest prominently in GPT3.5 V2, but were unlocked after the introduction of instruction tuning with RLHF \cite{chung2022scaling,ouyang2022training}.

Figures \ref{fig5}(c) and (d) respectively show line chart of the EM scores formed by each model in answering questions of different answer and reasoning types. The two Figures are consistent with what we observed in the Overall results, that is, various models of the GPT family and FLAN-T5 have similar line shapes. In addition, we also find that the performance of the new GPT models has improved significantly in some specific types of questions, such as Boolean-type (ChatGPT, GPT-4) and WHY-type (GPT-3.5 v3). However, in some other types of questions, there is no significant improvement for the multi-generation models of GPT, such as Num-type and Counting-type. This indicates that there is still a significant room for improvement for LLMs, and the iteration is far from over. Another interesting finding is that FLAN-T5 performs similarly to ChatGPT in answering boolean questions, but performs worse than the GPT family in other types of answers. Due to the difference in their training data, we cannot accurately determine in the current evaluation whether the reason for this situation is the difference in training data or whether certain training strategies used by the GPT family have a negative impact on specific types of questions.

\begin{table}\centering
\caption{MFT results of ChatGPT} \label{tab5}
\resizebox{\linewidth}{!}{
\setlength{\tabcolsep}{1.5mm}{
    \begin{tabular}{l|lllllll}
    \hline
     & SetOperation  & Filtering & Counting & Comparison & Single-hop & Multi-hop & Star-shape\\
    \hline
    Single Reasoning    & 60.22 & 51.39 & 24.16 & 31.48 & 44.07 & \textbf{48.27} & \textbf{50.75} \\
    Multiple Reasoning  & \textbf{70.00} & \textbf{63.40} & \textbf{28.41} & \textbf{74.74} & \textbf{54.00} & 44.88 & 47.43 \\
    \hline
    \end{tabular}
    }
    }
\end{table}

\begin{table}[t]\centering
\caption{INV results of GPT family}\label{tab6}
\resizebox{\linewidth}{!}{
\setlength{\tabcolsep}{0.75mm}{
    \begin{tabular}{l|ccccccccc}
    \hline
     LLM & CCC & CCW & CWC & CWW & WCC & WCW & WWC & WWW & Stability Rate  \\
     \hline
     GPT-3 & 434 & 64 & 59 & 52 & 42 & 43 & 73 & 666 & 76.76 \\
     GPT-3.5 v2 & 495 & 44 & 65 & 42 & 43 & 30 & 58 & 656 & 80.30\\
     GPT-3.5 v3 & 604 & 46 & 43 & 49 & 34 & 35 & 49 & 583 & 82.83 \\
     ChatGPT & 588 & 49 & 72 & 68 & 52 & 27 & 32 & 545 & 79.06 \\
     GPT-4 & 798 & 0 & 0 & 65 & 54 & 0 & 0 & 516 & \textbf{91.70}  \\
    % \multicolumn{3}{c}{Stability Rate}  & 79.06 & 79.44 \\
    \hline
    \end{tabular}
    }
    }
\end{table}

\begin{table}[t]\centering
\caption{DIR results for RsgType, the score represents the percentage of expected output produced by the LLMs. } \label{tab9}
% \resizebox{\linewidth}{!}{
\setlength{\tabcolsep}{1.5mm}{
    \begin{tabular}{lccccc}
    \hline
     &  SetOperation & Filtering & Counting & Comparison & Overall\\% & Failures example of test case & Expected \\
    \hline
    GPT-3.5 v3  & 45\% & 75\% & 65\% & \textbf{65\%} & 62.5\%\\
    ChatGPT     & \textbf{75\%} & 85\% & \textbf{70\%} & \textbf{65\%} & \textbf{73.75\%} \\
    GPT-4     & 65\% & \textbf{90\%} & \textbf{70\%} & 60\% & 71.25\% \\
    \hline
    \end{tabular}
    }
\end{table}

\subsection{CheckList results}
\textbf{MFT results}
In the MFT tests, we only evaluate questions that contain a single type of reasoning or multiple reasoning labels of the same type (such as SetOperation+Comparison and SetOperation+Filtering). Based on the results of MFT, we compared the performance of ChatGPT in answering single and multiple reasoning questions. Table \ref{tab5} shows the following findings. (1) Except for multi-hop and star type questions, ChatGPT performs better in executing multiple reasoning than in performing single reasoning in answering questions involving other types of reasoning operations. (2) ChatGPT is not good at answering counting questions despite the improvements generated by multiple reasoning.

% For the first finding, we speculate that one possible reason is that questions involving multiple types of reasoning provide richer context than questions involving only a single type of reasoning, which enhances ChatGPT’s ability to accurately identify target entities from its own knowledge. However, in multihop and star-type questions, which contain multiple known entities and at least one hidden entity (usually treated as an intermediate result of reasoning in traditional models), the increased context may be associated with the wrong entity, leading to negative effects on question answering. The second finding reveals that ChatGPT’s weakness in answering counting-type questions may be due to its text encoding rather than its lack of contextual learning. Our DIR results demonstrate that using prompt and chain-of-thought can significantly improve this issue. We also conducted an MFT test on other models in the GPT family, and the results can be found in the appendix.

\begin{table}[t]\centering
\caption{DIR results for AnsType prompting} \label{tab8}
% \resizebox{\linewidth}{!}{
\setlength{\tabcolsep}{1.2mm}{
    \begin{tabular}{lllllllll}
    \hline
     &  MISC & PER & LOC & ORG & DATE & Boolean & NUM & WHY\\
    \hline
    GPT-3      & \color{red}{+1.43} & 0 & \color{red}{+5.71} & \color{red}{+4.29} & \color{red}{+4.29} & \color{red}{+15.71} & \color{red}{+17.14} & 0 \\
    GPT-3.5 v2 & \color{blue}{-4.28}  & \color{red}{+2.85} & \color{red}{+7.14} & \color{red}{+14.28} & \color{red}{+2.86} & \color{blue}{-8.57} & \color{red}{+14.28} & \color{red}{+12.13}\\
    GPT-3.5 v3 & \color{blue}{-12.86} & \color{red}{+10.00} & \color{red}{+18.57} & \color{blue}{-7.14} & \color{red}{+4.71} & \color{red}{+17.14} & \color{red}{+22.85} & \color{red}{+9.09} \\
    ChatGPT    & \color{red}{+6.78}  & \color{blue}{-3.64} & \color{blue}{-1.72} & \color{blue}{-5.35} & \color{blue}{-8.58} & \color{red}{+4.28} & \color{red}{+7.15} & \color{blue}{-3.03} \\
    GPT-4      & \color{blue}{-4.29}  & \color{blue}{-2.86} & \color{red}{+11.43} & \color{red}{+5.71} & 0 & \color{red}{+7.14} & \color{red}{+4.29} & \color{blue}{-6.06} \\
    \hline
    \end{tabular}
    }
\end{table}

\textbf{INV results}
Table \ref{tab6} presents the stability of LLMs from the GPT family across three runs on three different test cases. As a reference, \cite{omar2023chatgpt} noted that the stability of traditional KBQA models is 100. The results in Table \ref{tab6} are reported using the following symbols: 'CCC' indicates that all answers to the three inquiries are correct, while 'WWW' indicates that none of the three inquiries received correct answers or the model did not return any useful answers. Only when the correctness of the three queries is consistent, the model's performance on the problem is considered stable. As shown in Tables \ref{tab6}, the overall stability of the GPT models has improved from GPT-3 to GPT-4, and GPT-4 has reached a stability rate of 91.70, which is very close to that of traditional KBQA models. The stability rate of ChatGPT is slightly lower than that of GPT-3.5, and we infer that this is due to the fact that the ChatGPT model focuses more on conversation training, resulting in higher instability (randomness) in the output.

\textbf{DIR results}
As mentioned in Figure \ref{fig4}, we designed three DIR modes to examine the controllability of LLMs from the GPT family. 
In the first mode, we manually observe whether the SPARQL statements output by the model contain the expected keywords, and calculate the failure rate of the model's expected reasoning operations. Since GPT-3.5 v2 and its earlier versions did not undergo code learning, it is difficult for them to generate correct SPARQL queries. Therefore, in this test, we compare the GPT-3.5 v3, ChatGPT, and GPT-4. 
% The results in Table \ref{tab9} show that the overall controllability of GPT-3.5 v3 for reasoning operations is weaker than that of the other two models, with ChatGPT performing slightly better than GPT-4. 
As shown in Table \ref{tab9}, the scores of around 73\% indicate that even the latest GPT model still has a high degree of randomness in performing reasoning operations, which will affect its applicable scenarios.

In the second mode, we provide prompts to the model's input indicating the answer type and observe the change in the EM score. In Table \ref{tab8}, red values indicate that adding prompts increases the EM score, while blue values indicate negative effects. For most models, prompts have a relatively stable positive effect on Boolean and NUM type questions, while the answers to MISC type questions are mostly negatively affected. In addition, in new models such as ChatGPT and GPT-4, the effect of answer type prompts is much worse than in GPT-3.5 and earlier models.This suggests that different models have different internal knowledge and understanding of the same input text, and the effectiveness and helpfulness of the same prompt vary among different models. More powerful models are more sensitive to the content of prompts because of their powerful natural language understanding ability. It may be difficult to design simple and universally effective prompts that work well across all models.

In the third mode, we guide the model step by step through a naive CoT-guided process to first provide the crucial information required to answer the question, and then answer the original question. Table \ref{tab10} shows the difference in EM scores of the GPT model's answers before and after using CoT-guided process for each type of questions. 
We can observe that positive impact brought by CoT to GPT-4 is greater than that of other models, and the improvement of CoT on the model's ability to answer NUM-type questions is significant and stable. In terms of reasoning types, CoT improves the ability of all models in set operations, conditional filtering, and counting, but it doesn't help much with multi-hop and star-shape questions. Specifically, the most significant improvement introduced by CoT for the GPT family of models is a score increase of over 30.00 for answer types that are numerical(NUM). This result strongly supports the importance of CoT for using LLMs to solve numerical-related questions \cite{kojima2022large}.

\begin{table}[t]\centering
\caption{DIR results for CoT prompting} \label{tab10}
\resizebox{\linewidth}{!}{
\setlength{\tabcolsep}{1.2mm}{
    \begin{tabular}{lcccccccc}
    \hline
     &  MISC & PER & LOC & ORG & DATE & Boolean & NUM & WHY\\
    \hline
    GPT-3      & \color{blue}{-1.40}  & \color{blue}{-2.00} & \color{blue}{-2.67} & \color{red}{+2.73} & \color{blue}{-3.77} & \color{red}{+3.36} & \color{red}{+35.66} & \color{red}{+6.06} \\
    GPT-3.5 v2 & \color{blue}{-0.35} & \color{blue}{-5.33} & \color{red}{+1.78} & \color{blue}{-3.64} & \color{red}{+0.76} & \color{blue}{-5.04} & \color{red}{+32.95} & 0 \\
    GPT-3.5 v3 & 0 & \color{blue}{-2.00} & \color{blue}{-1.33} & \color{blue}{-1.82} & \color{blue}{-1.51} & \color{blue}{-2.10} & \color{red}{+34.12} & 0 \\
    ChatGPT    & \color{blue}{-1.75}  & \color{blue}{-4.66} & \color{red}{+0.89} & \color{blue}{-3.63} & \color{blue}{-1.50} & \color{red}{+3.36} & \color{red}{+30.62} & \color{red}{+6.06} \\
    GPT-4      & \color{blue}{-3.00} & \color{red}{+11.11} & \color{red}{+2.22} & \color{red}{+3.3} & \color{blue}{-2.71} & 0 & \color{red}{+20.00} & \color{red}{+2.62} \\
    \hline
     &  SetOperation & Filtering & Counting & Comparison & Multi-hop & Star-shape &  & \\
    \hline
    GPT-3      & \color{red}{+10.79} & \color{red}{+10.43} & \color{red}{+35.66} & \color{red}{+1.35} & \color{blue}{-1.60} & \color{blue}{-1.69} &  &  \\
    GPT-3.5 v2 & \color{red}{+4.86} & \color{red}{+5.46} & \color{red}{+38.54} & \color{blue}{-2.26} & \color{blue}{-1.18} & \color{blue}{-0.85} &  &  \\
    GPT-3.5 v3 & \color{red}{+6.34} & \color{red}{+8.18} & \color{red}{+38.99} & \color{blue}{-1.13} & \color{blue}{-1.61} & \color{blue}{-1.26} &  &   \\
    ChatGPT    & \color{red}{+7.82}  & \color{red}{+9.47} & \color{red}{+35.78} & \color{red}{+0.45} & \color{blue}{-1.47} & \color{blue}{-1.41} &  & \\
    GPT-4      & \color{red}{+2.05} & \color{red}{+0.93} & \color{red}{+11.11} & \color{blue}{-1.88} & \color{red}{+2.82} & \color{red}{+2.68} &  &  \\
    \hline
    \end{tabular}
    }
    }
\end{table}

% \begin{table}\centering
% \caption{DIR results for CoT prompting} \label{tab10}
% % \resizebox{\linewidth}{!}{
% \setlength{\tabcolsep}{1.2mm}{
%     \begin{tabular}{c|c|c|c}
%     \hline
%      Types &  Without prompting & With prompting & +/- \\
%     \hline
%     MISC    & 64.33 & 62.58 & -1.75 \\
%     PER     & 53.33 & 48.67 & -4.66 \\
%     LOC     & 63.11 & 64.00 & +0.89 \\
%     ORG     & 68.18 & 64.55 & -3.63 \\
%     DATE    & 31.57 & 30.07 & -1.50 \\
%     Boolean & 75.63 & 78.99 & +3.36 \\
%     NUM     & 23.64 & 54.26 & +30.62 \\
%     WHY     & 39.39 & 45.45 & +6.06 \\
%     UNA     & - & - & - \\
%     \hline
%     SetOperation   & 83.51  & 91.33  & +7.82 \\
%     Filtering   & 69.50  & 78.97  & +9.47 \\
%     Counting    & 27.52  & 63.30  & +35.78 \\
%     Comparison  & 86.00  & 86.45  & +0.45 \\
%     Multi-hop   & 43.11  & 41.64  & -1.47 \\
%     Star-shape  & 43.44  & 42.03  & -1.41 \\
%     \hline
%     \end{tabular}
%     }
% \end{table}

\section{Conclusion}
In this paper, we extensively tested the ability of ChatGPT and other LLMs to answer questions on KB-based CQA datasets using their own knowledge. 
The experimental results showed that the question-answering performance and reliability of the GPT model have been continuously improving with version iterations, approaching that of traditional models. CheckList testing showed that current LLMs still have a lot of room for improvement in some reasoning abilities, and CoT-inspired prompts can improve the performance of the original model on certain types of questions. Consequently, this evaluation serves as a valuable reference for future research in the relevant community.
% We believe that this evaluation helps to understand the strengths and limitations of LLMs in KB-based CQA tasks. It also assists in analyzing the ability of LLMs to organize and call upon their inherent knowledge, and to explore the commonalities between the GPT family and other prevalent open-source language models. Consequently, this evaluation serves as a valuable reference for future research in the relevant community.
In future work, we need to further expand on the following two points: Firstly, conduct tests in various domains to validate which conclusions obtained from open-domain KBQA are universal and which are domain-specific. Secondly, perform tests on various types of models. With ongoing LLM research, besides the GPT family, many new large-scale open-source models have been proposed. It requires further exploration and summarization to determine if they possess better mechanisms for self-knowledge organization or stronger capabilities to accept human prompts and find answers.
% 在未来的工作中，我们还需就以下两点作进一步拓展：其一，面向更多领域的测试，以验证开放域KBQA上得到的结论哪些是普遍存在的，而哪些是领域特有的；其二，面向更多模型种类的测试，随着LLM研究的继续，除GPT家族外，许多新的开源大模型被提出，它们是否具备更好的自身知识组织的机制或更强的接受人类提示以寻找答案的能力，都需要进一步的探索和归纳。

\section{Acknowledgments}
This work is supported by the Natural Science Foundation of China (Grant No.U21A20488). We thank the Big Data Computing Center of Southeast University for providing the facility support on the numerical calculations in this paper.
% In future work, we

% This paper presents a large-scale experimental analysis of ChatGPT’s performance in answering complex questions using its own knowledge base compared to similar models and current state-of-the-art (SOTA) models. The analysis highlights the advantages, limitations, and deficiencies of ChatGPT. Using a checklist test, we provide a detailed analysis of ChatGPT’s basic performance, stability, and controllability in dealing with various answer types and reasoning requirements. We believe that these findings will provide valuable insights and references for the development and downstream research of large-scale language models represented by ChatGPT.

\bibliographystyle{splncs04}
\bibliography{splncs-refs}




% LNCS Homepage, \url{http://www.springer.com/lncs}. Last accessed 4
% \end{thebibliography}
\end{document}